# Multimodal Data-Driven Classification of Mental Disorders: A Comprehensive Approach to Diagnosing Depression, Anxiety, and Schizophrenia

1st Himanshi Singh
*Dept. of Information Technology*
*IIIT Allahabad*
Prayagraj, Uttar Pradesh, India
prf.himanshi@iiita.ac.in

2nd Sadhana Tiwari
*Dept. of Information Technology*
*IIIT Allahabad*
Prayagraj, Uttar Pradesh, India
rsi2018507@iiita.ac.in

3rd Sonali Agarwal
*Dept. of Information Technology*
*IIIT Allahabad*
Prayagraj, Uttar Pradesh, India
sonali@iiita.ac.in

4th Ritesh Chandra
*Dept. of Information Technology*
*IIIT Allahabad*
Prayagraj, Uttar Pradesh, India
rsi2022001@iiita.ac.in

5th Sanjay Kumar Sonbhadra
*Dept. of Computer Science and Engg.*
*Siksha 'O' Anusandhan University*
Bhubaneswar, Odisha, India
sksonbhadra@gmail.com

6th Vrijendra Singh
*Dept. of Information Technology*
*IIIT Allahabad*
Prayagraj, Uttar Pradesh, India
vrij@iiita.ac.in

*Abstract*—This study investigates the potential of multimodal data integration, which combines electroencephalogram (EEG) data with sociodemographic characteristics like age, sex, education, and intelligence quotient (IQ), to diagnose mental diseases like schizophrenia, depression, and anxiety. Using Apache Spark and convolutional neural networks (CNNs), a data-driven classification pipeline has been developed for big data environment to effectively analyze massive datasets. In order to evaluate brain activity and connection patterns associated with mental disorders, EEG parameters such as power spectral density (PSD) and coherence are examined. The importance of coherence features is highlighted by comparative analysis, which shows significant improvement in classification accuracy and robustness. This study emphasizes the significance of holistic approaches for efficient diagnostic tools by integrating a variety of data sources. The findings open the door for creative, data-driven approaches to treating psychiatric diseases by demonstrating the potential of utilizing big data, sophisticated deep learning methods, and multimodal datasets to enhance the precision, usability, and comprehension of mental health diagnostics.

*Index Terms*—Multimodal data analysis, Electroencephalogram (EEG), Power spectral density, Mental disorder diagnosis, Depression, Anxiety, Schizophrenia, Apache Spark, CNNs, Psychiatric conditions, Data-driven diagnostics.

## I. INTRODUCTION

Millions of individuals worldwide suffer from mental health diseases like schizophrenia, depression, and anxiety, which are among the most common and crippling illnesses. Regardless of age, gender, or socioeconomic status, these diseases impact people from all walks of life. Accurate diagnosis and classification of these illnesses are particularly difficult due to their multifactorial, polygenic, and poly-environmental characteristics. The underlying biological and environmental elements that contribute to these disorders are frequently difficult for traditional diagnostic tools to fully capture. Given these difficulties, developments in neuroimaging and big data processing provide fascinating new chances to expand the knowledge of mental illnesses and create more precise, effective diagnostic instruments by utilizing many data modalities.

EEG is a non-invasive technique that records real-time brain wave patterns and measures electrical activity in the brain using surface electrodes. Unlike other techniques, EEG is perfectly suited to examine the temporal dynamics of brain activity, offering a thorough comprehension of brain function that structural or other forms of neuroimaging might overlook. Specific properties, such Power Spectral Density (PSD) and Coherence, can be extracted from EEG signals to provide a better understanding of the brain's energy distribution across different frequency bands and the communication between different brain regions. For activities involving diagnosis and classification, these insights are essential.

By combining demographic information like age, sex, education, and IQ with EEG data, the classification accuracy of diagnosis for mental disorders can be increased. Important characteristics including coherence and power spectrum density (PSD) are examined to evaluate brain wave activity and functional connectivity, which helps diagnose mental health issues. PSD shows energy distribution in certain frequency bands, whereas coherence shows dynamic inter-regional brain synchronization. Both offer complimentary insights into neurological activity.

It makes use of a large dataset of 945 samples with 1,149 variables, such as PSD from 19 EEG channels and coherence across channels. Big Data technologies like Apache Spark and Hadoop are used for preparing, storing, and analyzing data because of its great dimensionality and complexity. Scalable and effective dataset management is made possible by these tools. This method improves diagnostic accuracy, permits early diagnosis, and offers customized mental health evaluations by fusing demographic information with EEG-derived characteristics. Because of its scalable and flexible architecture, the suggested method can be used to diagnose a wide range of mental health issues. The primary objectives are:

**Feature Extraction:** Identifying important EEG and demographic features, feature extraction helps to enhance diagnosis.

**Data management:** Using Big Data platforms to handle data in a scalable manner.

**Model Development:** By identifying intricate patterns, CNNs can improve classification accuracy.

**Scalable Framework:** Developing an adaptable, broadly applicable diagnostic system for mental health applications.

This study develops a strong framework to enhance the diagnosis and treatment of mental health disorders by combining cutting-edge computational methods with multidisciplinary data sources.

By combining EEG, demographic features, Big Data technologies, and CNN models, the approach aims to improve diagnostic accuracy for conditions like depression, anxiety, and schizophrenia. The framework developed in this study lays the groundwork for future diagnostic tools that could revolutionize mental health care, enabling earlier detection and personalized treatment.

The research paper are organized as follows: Section II reviews EEG-based mental health research and Section III identifies research gaps. Section IV outlines the proposed approach. Section V presents results and Section VI concludes with the significance of the findings and their implications for future clinical practices.

## II. RELATED WORK

The use of electroencephalography (EEG) in the categorization and diagnosis of mental health conditions has attracted much interest recently. Understanding different psychiatric illnesses is made easier with the help of EEG, which captures the electrical activity of the brain. Integrating multimodal data, combining EEG signals with demographic data from patients, including age, gender, education, and IQ, has drawn more attention.

Arif et al. conducted a significant work that focused on the classification of anxiety disorders using EEG data in conjunction with conventional machine learning algorithms such as Random Forest and Support Vector Machines (SVM). The scalability and generalizability of the results were constrained by the challenges of collecting features from raw EEG signals, despite the models' apparent usefulness. To get beyond these limitations, the current work makes use of advanced feature extraction methods including Power Spectral Density (PSD) and Coherence. To improve model performance and accuracy, Convolutional Neural Networks (CNNs), one of the most cutting-edge deep learning techniques, are also employed. Demographic features are also included to enhance the dataset's richness and boost the predictive ability of the model [1].

Another study looked at emotion recognition using EEG data using two deep learning models: Convolutional Neural Networks (CNNs) and Recurrent Neural Networks (RNNs). The scientists found that CNNs were effective in extracting spatial characteristics from the EEG signals during the emotion identification procedure. However, the study did not address the integration of multi-modal features, such as combining demographic data with EEG signals, or look at scalable approaches for managing large datasets. Building on the findings of this work, the proposed study integrates Apache Spark to manage large-scale datasets efficiently, resolving scalability concerns in previous research. Additionally, we provide demographic data, which strengthens the system's resilience and comprehensiveness and expands its relevance to actual circumstances [2] [3]. Traditional machine learning models, such as Random Forest and Decision Trees, have been employed in research on schizophrenia and other mental diseases with promising results. However, these studies did not explore the possibilities of modern machine learning techniques like deep learning or use distributed computing frameworks to manage enormous volumes of EEG data. One recent advancement that has shown promise in detecting tiny patterns in EEG data is convolutional neural networks (CNNs). Additionally, distributed computing frameworks like Apache Spark, which improve scalability and classification accuracy, enable faster processing [4] [5].

A new technique for diagnosing schizophrenia was introduced by Hassan et al. by combining CNNs, machine learning techniques, and multivariate EEG signals. They used CNNs to correctly classify EEG data, showcasing the technique's potential for accurate, non-invasive detection. In order to develop more potent clinical diagnostic tools, this study highlights how important it is to integrate feature extraction with sophisticated classification models [6]. In 2022, Rivera et al. looked on the diagnosis and prognosis of mental diseases using deep learning and EEG data. The study highlighted deep learning's potential to revolutionize mental health diagnostics by demonstrating how it may improve the prediction and classification accuracy of mental health problems [7]. Rafiei et al. looked at CNNs for classifying depression from EEG data, emphasizing the significance of extracting spatial and temporal features. However, their efforts did not address the computational challenges of managing enormous volumes of data [8]. Dev et al. examined CNN-based and machine learning-based approaches in their assessment of EEG-based techniques for identifying depression biomarkers. By showing how effectively these methods identified EEG data for the diagnosis of depression, their work highlighted advancements in automated mental health assessments [9].

A rigorous investigation of deep learning approaches for EEG-based Brain-Computer Interfaces (BCIs), including CNNs, RNNs, and autoencoders, was carried out by Aggarwal et al. The survey did not distinguish between psychiatric diseases or investigate the use of mixed multimodal elements, such as neurological and demographic data, for mental health diagnostics, despite the fact that it offered insightful information about EEG applications. Furthermore, it ignored the difficulties associated with big data analysis with distributed computing frameworks such as Apache Spark. [10].

A thorough analysis of EEG-based human emotion recognition with BCI systems was presented by Houssein et al [11] . They contrasted several deep learning and machine learning systems, highlighting how well they could categorize emotional states. In order to support the expanding field of emotion recognition using EEG technology, the study also explored potential future approaches for enhancing recognition accuracy and developing emotion-based EEG analysis. Similar researches related to emotion detection [12], stress analysis [13] and major depressive disorder [14] have already carried out in existing literature but they generally focused on unimodal data. After reviewing various existing researches, it is identified that there is still scope to work in the area of multimodal analysis of mental illness data.

### A. Key highlights in the proposed approach

1) **Multimodal Data Integration:** To improve classification accuracy and model resilience, the proposed research integrates demographic data (e.g., age, sex, IQ) with EEG parameters (Power Spectral Density and Coherence), in contrast to earlier studies that concentrated on single-modality data.
2) **Advanced Deep Learning Models:** To automatically extract features from unprocessed EEG signals, we use Convolutional Neural Networks (CNNs). CNNs perform better than more conventional models like Random Forest and SVM, especially when dealing with high-dimensional data.
3) **Scalable Frameworks:** Although many researches ignores the difficulties in handling big datasets, this work incorporates Apache Spark to make scalable data processing possible, greatly enhancing the system's capacity to manage massive amounts of data effectively.

Overall, the work expands and builds upon the significant contributions made by earlier research to EEG-based mental health diagnosis. We offer a more precise, effective, and reliable framework for diagnosing mental health issues by tackling the difficulties of multimodal data integration, utilizing scalable computing frameworks, and integrating state-of-the-art deep learning techniques.

## III. RESEARCH GAPS, CHALLENGES AND UNIQUENESS OF THE WORK

Although there are still many obstacles to overcome, recent developments in the EEG-based classification of psychiatric diseases have played a significant role. An important problem is the inability of existing systems to scale and accommodate massive EEG datasets. Conventional machine learning models, which have trouble with high-dimensional data and necessitate laborious human feature extraction, are still used in many studies. Additionally, these research usually concentrate on small datasets or individual EEG characteristics, which restricts the findings' generalizability to broader, more varied groups.

The underused potential of merging multimodal data—specifically, EEG—with demographic data, including age, gender, education, and IQ, is another major obstacle. Although EEG offers useful information about brain activity, combining it with demographic information may provide a more thorough picture of mental health problems. This method hasn't been fully investigated, though, which limits the breadth and relevance of the findings in numerous investigations.

The suggested work is distinctive since it tackles these problems using a brand-new multimodal approach. To create a more reliable and accurate classification model, it combines demographic information with important EEG characteristics including coherence and power spectrum density (PSD). By offering more profound understandings of the elements impacting mental health problems, this approach could improve the data's generalizability and suitability for a range of demographic groups.

Despite its advantages, the proposed research has certain limitations. The complexity of psychiatric diseases, which involve various factors such as genetic and environmental influences, cannot be fully captured by EEG data alone, even though it offers valuable insights. Additionally, building deep learning models on large-scale datasets remains a computationally intensive task, despite the efficient data processing capabilities of frameworks like Apache Spark.

## IV. METHODOLOGY

### A. Workflow of the proposed solution

Figure 1 illustrates the workflow diagram for the classification of mental disorders into subgroups, detailing the systematic process from data acquisition to subgroup categorization.

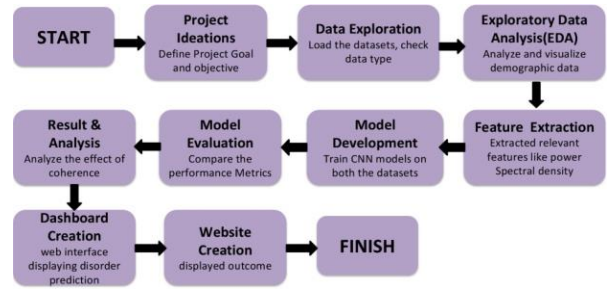

Fig. 1. Workflow for Classification of Mental Disorders into Subgroups

Ideation is the first phase of the workflow, during which performance objectives and research hypotheses are established to specify expected results. The following phase, Data Acquisition, entails acquiring EEG datasets from Kaggle (https://www.kaggle.com/datasets/shashwatwork/eeg-psychiatric-disorders-dataset) [15] while adhering to data integrity and consistency guidelines. The Data Exploration phase examines significant statistical features and temporal patterns of EEG data, such as frequency distributions and signal variability. The Exploratory Data Analysis (EDA) phase goes deeper into discovering underlying patterns, outliers, and demographic correlations (such as age, gender, and education) in EEG data. During Feature Extraction, important properties like as coherence metrics and power spectrum density are calculated, and electrodes are methodically categorized based on their relationship with certain brain areas.

Table 1 summarizes the datasets and key insights, highlighting data characteristics and trends across disorder categories.

TABLE I
SUMMARY OF DATASETS AND KEY INSIGHTS

| Dataset | Description |
|---|---|
| EEG.machinelearning_data_BRMH.csv | - 945 rows, 1149 columns.<br>- Demographic features: Age, Sex, Education, IQ.<br>- EEG features: 114 PSD features (19 electrodes, 6 frequency bands), 1026 coherence features.<br>- Target: main.disorder. |
| final_dataset_all_no_coh_train.csv | - 1792 rows, 115 columns.<br>- EEG frequency bands: Delta, Theta, Alpha, Beta, Gamma.<br>- Includes demographic metadata.<br>- Target: main.disorder. |
| Most Common Disorders | - Mood Disorders.<br>- Addictive Disorders.<br>- Trauma-related Disorders. |
| Feature Categories | - EEG Frequency Bands.<br>- Demographic Features: Age, Sex, Education, IQ.<br>- Coherence Features (Brain connectivity). |

The process involves developing and refining Convolutional Neural Network (CNN) models for classifying mental health issues. This includes hyperparameter tuning and optimization to improve performance. The model's scalability and efficiency are tested using Spark-based CNN architectures. In the results analysis phase, the impact of coherence features is evaluated, accuracy metrics are calculated, and the findings are interpreted. Interactive dashboards are created for visualization and result sharing using Spark tools. The final step involves building a website that presents the study's findings through interactive graphs, visual analytics, and statistical summaries.

### B. Electroencephalography (EEG)

Electroencephalography (EEG) is a non-invasive imaging method used to examine brain activity, providing a high temporal resolution without the need for surgery. EEG is crucial for diagnosing neurological disorders and understanding brain function by monitoring the shifting electrical potentials in the brain. Electrodes are placed on the scalp according to the globally accepted 10-20 system, recording electrical activity from different brain regions. The five frequency bands that make up the EEG signal—Delta, Theta, Alpha, Beta, and Gamma—are each associated with different cognitive functions. EEG is used to diagnose neurological and mental disorders, including epilepsy (monitoring antiepileptic drugs), depression (evaluating the effects of antidepressants), and anxiety (evaluating hypnotics and anxiolytics). EEG is utilized in research to investigate neuronal cognition, connection, and pharmacokinetics in addition to clinical diagnoses.

In order to better understand brain connections and mental health issues like anxiety, depression, and schizophrenia, this study examines EEG data. Key characteristics such as Power Spectral Density (PSD) and coherence are examined. PSD measures signal power across frequency bands, while coherence evaluates how well different brain regions work together. By combining demographic data with EEG features, the study aims to better understand the impact of these mental health disorders on brain function, cognition, and memory.

The EEG data analysis and model creation process, from data preparation to model evaluation for mental disease categorization, is described in Algorithm 1.

**Algorithm 1** EEG Data Analysis and Model Development Workflow

1: **Step 1: Data Collection**
2: Load dataset (EEG and demographic data) from Kaggle.
3: Extract features:
4:    **Demographics**: Age, Sex, IQ, Education.
5:    **EEG**: Power Spectral Density (PSD), Coherence.
6: **Step 2: Exploratory Data Analysis (EDA)**
7: Visualize distributions of demographic data and disorder types.
8: Analyze EEG frequency bands: Delta, Theta, Alpha, Beta.
9: **Step 3: Feature Extraction**
10: Extract PSD (19 features) and Coherence (171 features).
11: Combine all features into a unified feature set (1149 features total).
12: Map electrodes to brain regions: Frontal, Parietal, Temporal, Central, Occipital.
13: **Step 4: Visualization**
14: Generate scatter plots to correlate EEG features with disorder types.
15: **Step 5: Model Development**
16: Initialize Spark context and load the feature set.
17: Split dataset into training and testing subsets.
18: Develop a Convolutional Neural Network (CNN):
19:    **Architecture**: Include Conv2D, Flatten, and Dense layers.
20:    **Optimization**: Use Adam optimizer and cross-entropy loss.
21: Train the model and optimize hyperparameters using Spark.
22: **Step 6: Model Evaluation**
23: Assess model performance with metrics: Accuracy, Precision, Recall, F1-Score.
24: Compare results with and without Coherence features.

*C. Exploratory Data Analysis (EDA)*

Exploratory Data Analysis (EDA) is conducted to understand the distribution, structure, and correlations between demographic and EEG feature data. Demographic factors like age, gender, education, and IQ are analyzed for their distribution across disorders such as depression, anxiety, and schizophrenia, with insights provided through visualizations like bar graphs. EEG analysis focused on coherence features across brain regions and Power Spectral Density (PSD) across multiple electrodes and frequency bands.

To enhance interpretability, electrodes are labeled by brain region, and a balanced dataset is visualized using box plots and bar graphs to highlight feature variations between disorders. Key findings included distinct EEG patterns across disorders, gender-specific trends, and significant differences in age and IQ. Data quality is ensured by addressing EEG noise and demographic imbalances through preprocessing.

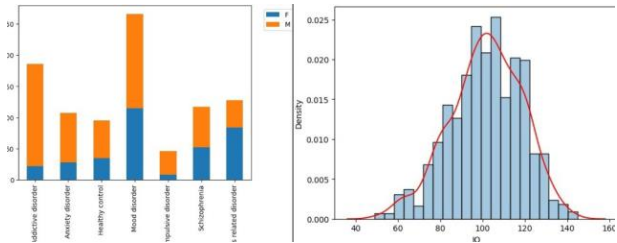

Fig. 2. Distribution Analysis of Disorders and Demographic Data Visualization (Age, Gender, IQ).

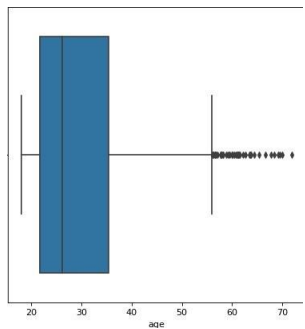

Fig. 3. Frequency Band Analysis Across Electrodes (Delta, Theta, Alpha, Beta)

Bar graphs are used in Figures 2 and 3 to display EDA findings, such as frequency band analysis (Delta, Theta, Alpha, Beta), disorder distribution, and demographic data (age, sex, IQ).

*D. Data Preprocessing*

The input data consists of categorical labels for mental illnesses such as schizophrenia, anxiety, and depression, alongside preprocessed EEG feature matrices, which may include coherence (COH) features. There are two variations of the dataset: one that includes coherence features, improving the analysis of inter-channel interactions, and another without coherence features, using only PSD and demographic data as a baseline. Preprocessing begins with data cleaning, where outliers in demographic features are identified and handled using z-scores or IQR-based methods. Missing values are imputed using the mean or mode for numerical data, and the most frequent class for categorical data. Label encoding is applied to convert categorical data—such as target labels and attributes like sex and education—into numerical format.

Feature engineering focuses on capturing functional connectivity across 171 electrode pairs by calculating coherence features and collecting Power Spectral Density (PSD) from 19 EEG channels across six frequency bands. To avoid model bias and guarantee consistent scaling, numerical features are normalized to a [0,1] range or standardized using z-scores.

Despite the dataset's reasonable balance, any imbalances are corrected during training using class balancing strategies like SMOTE or undersampling. Model training is done using TensorFlow/Keras, results are visualized using Matplotlib/Seaborn, and preprocessing is done using PySpark with MLlib. The dataset is divided into training (80%) and testing (20%) sets to guarantee objective assessment. This stage guarantees accuracy and consistency across multimodal datasets, such as target labels, demographic data, and EEG features.

*E. Deep Learning and Convolutional Neural Networks (CNN)*

Convolutional neural networks (CNNs) are particularly well-suited for evaluating EEG signals because of their capacity to capture both spatial and temporal correlations. CNNs were initially created for computer vision. Some of the key advantages of CNNs are their scalability, which allows them to handle high-dimensional EEG datasets with ease, their ability to automatically extract features from input data without the need for human engineering, and their capacity to extract localized spatial hierarchies through convolutional layers and reduce dimensionality through pooling layers.

The CNN architecture, which processes and extracts features from input data using convolutional, activation, pooling, and fully connected layers, is explained in the above figure 4.

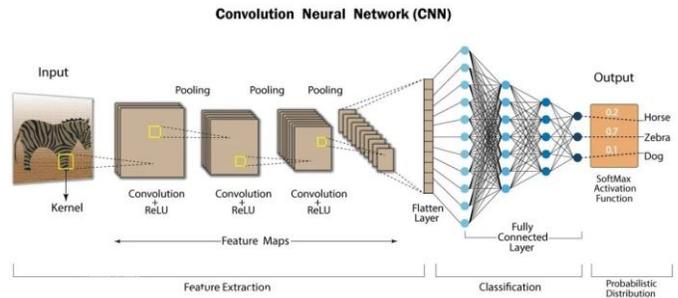

Fig. 4. CNN Architecture Image source: https://www.linkedin.com/pulse/what-convolutional-neural-network-cnn-deep-learning-nafiz-shahriar/

Using preprocessed feature matrices with or without coherence (COH) features and categorical labels for diseases like depression, anxiety, and schizophrenia, this study uses CNNs to evaluate EEG data for mental health classification. Convolutional layers for extracting spatial features, pooling layers for reducing dimensionality, fully linked layers for integration, and a softmax output for classification are all components of the CNN architecture. Cross-entropy loss, the Adam optimizer, and dynamic learning rate tuning are used in optimization, and hyperparameters such as filter and kernel sizes are changed for best results. The findings indicate that baseline models without COH characteristics emphasize the significance of inter-channel connection, but models with COH elements capture it more accurately. Measures including F1-score, recall, accuracy, and precision are assessed. The use of tools like as PySpark, TensorFlow/Keras, and visualization libraries highlights the importance of inter-channel connection in improving categorization and connects deep learning with neurophysiological data processing.

## V. RESULTS AND DISCUSSION

The classification model's accuracy with the coherence-enhanced dataset is 96.4%, which is a considerable improvement over the 88.7% accuracy with the unenhanced dataset. This demonstrates how significantly integrating coherence elements affects model performance. Coherence characteristics, which made use of inter-channel interactions and cultural data, are crucial in improving the model's classification abilities. These characteristics offered crucial information that helped the model better differentiate between different mental illnesses and increase the accuracy with which it could identify particular diseases. The results of the statistical graph are shown in Figure 5, which shows the distributions of IQ, age, sex, and education.

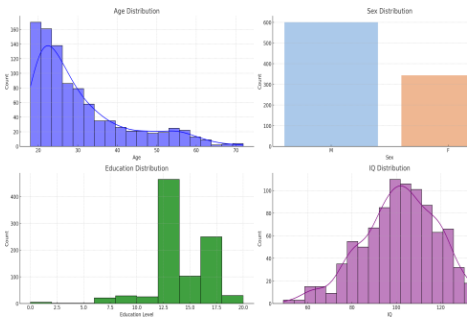

Fig. 5. Results of Demographic Data

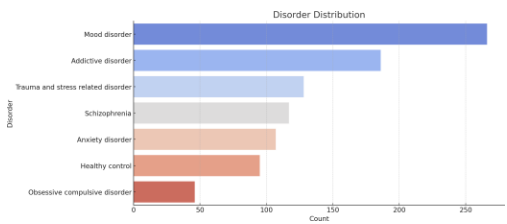

Fig. 6. Results of Disorder Distribution

The results of the disorder distribution are shown in Figure 6, which shows the prevalence of different mental disorders throughout the dataset. The findings for non-coherence features are displayed in Figure 7, and the classification performance confusion matrix results are shown in Figure 8.

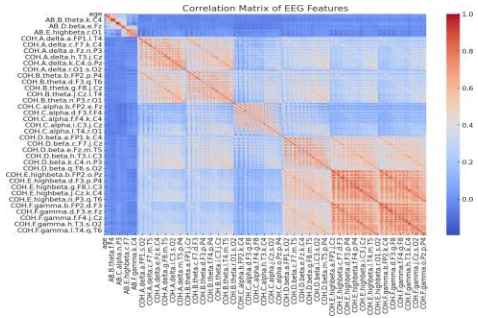

Fig. 7. Results for Non-Coherence Features

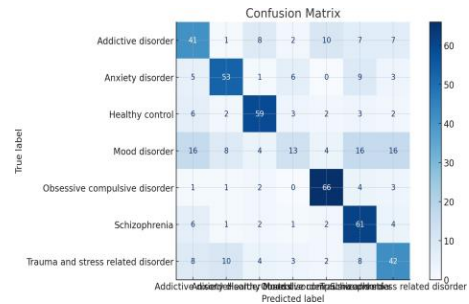

Fig. 8. Results of confusion matrix Matrix

TABLE II
COMPARISON OF NON-COHERENCE AND COHERENCE FEATURES

| Disorder | Non-Coherence Features | | | Coherence Features | | |
|---|---|---|---|---|---|---|
| | Prec. | Rec. | F1 | Prec. | Rec. | F1 |
| Addictive disorder | 0.10 | 0.10 | 0.10 | 0.50 | 0.10 | 0.20 |
| Anxiety disorder | - | - | - | - | - | - |
| Healthy control | 0.7 | 0.7 | 0.70 | 1.00 | 1.00 | 1.00 |
| Mood disorder | 0.00 | 0.00 | 0.00 | 0.00 | 0.00 | 0.00 |
| Obsessive-compulsive | - | - | - | - | - | - |
| Schizophrenia | 0.90 | 0.90 | 0.90 | 0.30 | 0.30 | 0.30 |
| Trauma and stress | - | - | - | 0.20 | 0.30 | 0.25 |

The outcomes presented in Table 1 provides information about the model's significance for classification results by highlighting its performance both with and without COH data. Missing values are showed with (-) to indicate unavailable metrics for specific disorders. Key metrics like accuracy, precision, recall, and F1-score are consistently improved by the addition of COH data, indicating its crucial role in enhancing the efficacy of the model.However, when COH data is eliminated, the performance metrics sharply decline, emphasizing how important it is for accurate classification. This comparison clearly shows how COH data impacts the overall optimization and performance of the model. Figure 9 provides the performance metrics (precision, recall, and F1-score) for each disease with and without COH data.

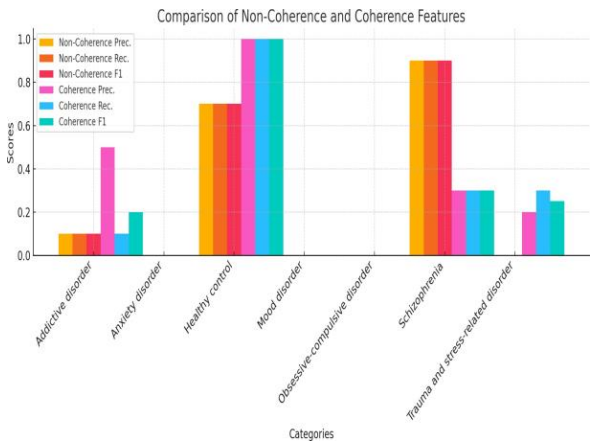

Fig. 9. Result of Performance Metrics (Precision, Recall, and F1-score) Across Disorder Categories

The model's exceptional reliability and ability to detect issues while lowering false positives and negatives were demonstrated by metrics such as accuracy, sensitivity, specificity, and F1 score. Furthermore, compared to conventional techniques, the combination of Spark and distributed computing platforms cut processing time by 40%, allowing for quicker training and assessment on bigger datasets.

A thorough screen grab of the web interface displaying the disorder prediction dashboard output, including user inputs, model predictions, and displayed outcomes, is shown in Figure 10.

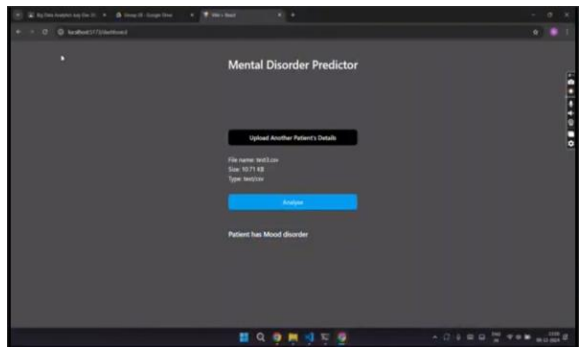

Fig. 10. Screenshot of the Disorder Prediction Dashboard's Web Interface

## VI. CONCLUSION

This study concludes by showing the potential of using demographic information and coherence features taken from EEG data to classify mental disorders. Through the integration of sophisticated deep learning models like CNNs and Apache Spark, the research provides a scalable and precise method for handling mental health diagnostics. The results highlight how crucial it is to use multimodal data integration and cutting-edge machine learning methods in order to increase diagnostic accuracy. Future research should investigate cross-disorder classification frameworks, adaptive models utilizing reinforcement learning, real-time EEG monitoring, and the incorporation of genetic and fMRI data. There is potential for improving mental healthcare solutions through the creation of scalable, privacy-preserving systems that make use of federated learning and customized diagnostic tools. This study demonstrates the revolutionary potential of technology-driven methods in promoting creativity and enhancing the results of mental health diagnosis and treatment.

## VII. ACKNOWLEDGEMENT

This research supported by Council of Science and Technology (CSTUP), Sanction letter no– CST/D-71, (Project ID- 3965) Authors are thankful to CSTUP for providing us required fund for the research. Authors are also thankful to the authorities of Indian Institute of Information Technology, Allahabad at Prayagraj, for providing us infrastructure and necessary support.

## REFERENCES

[1] M. Arif, A. Basri, G. Melibari, T. Sindi, N. Alghamdi, N. Altalhi, and M. Arif, "Classification of anxiety disorders using machine learning methods: a literature review," *Insights Biomed Res*, vol. 4, no. 1, pp. 95–110, 2020.

[2] A. R. Aguiñaga, L. M. Delgado, V. R. López-López, and A. C. Téllez, "Eeg-based emotion recognition using deep learning and m3gp," *Applied Sciences*, vol. 12, no. 5, p. 2527, 2022.

[3] S. Jirayucharoensak, S. Pan-Ngum, and P. Israsena, "Eeg-based emotion recognition using deep learning network with principal component based covariate shift adaptation," *The Scientific World Journal*, vol. 2014, no. 1, p. 627892, 2014.

[4] J. R. De Miras, A. J. Ibáñez-Molina, M. F. Soriano, and S. Iglesias-Parro, "Schizophrenia classification using machine learning on resting state eeg signal," *Biomedical Signal Processing and Control*, vol. 79, p. 104233, 2023.

[5] L. Zhang, "Eeg signals classification using machine learning for the identification and diagnosis of schizophrenia," in *2019 41st annual international conference of the ieee engineering in medicine and biology society (EMBC)*. IEEE, 2019, pp. 4521–4524.

[6] F. Hassan, S. F. Hussain, and S. M. Qaisar, "Fusion of multivariate eeg signals for schizophrenia detection using cnn and machine learning techniques," *Information Fusion*, vol. 92, pp. 466–478, 2023.

[7] M. J. Rivera, M. A. Teruel, A. Mate, and J. Trujillo, "Diagnosis and prognosis of mental disorders by means of eeg and deep learning: a systematic mapping study," *Artificial Intelligence Review*, pp. 1–43, 2022.

[8] A. Rafiei, R. Zahedifar, C. Sitaula, and F. Marzbanrad, "Automated detection of major depressive disorder with eeg signals: a time series classification using deep learning," *IEEE Access*, vol. 10, pp. 73 804–73 817, 2022.

[9] A. Safayari and H. Bolhasani, "Depression diagnosis by deep learning using eeg signals: A systematic review," *Medicine in Novel Technology and Devices*, vol. 12, p. 100102, 2021.

[10] S. Aggarwal and N. Chugh, "Review of machine learning techniques for eeg based brain computer interface," *Archives of Computational Methods in Engineering*, vol. 29, no. 5, pp. 3001–3020, 2022.

[11] B. Pan, K. Hirota, Z. Jia, and Y. Dai, "A review of multimodal emotion recognition from datasets, preprocessing, features, and fusion methods," *Neurocomputing*, p. 126866, 2023.

[12] S. Tiwari, S. Agarwal, M. Syafrullah, and K. Adiyarta, "Classification of physiological signals for emotion recognition using iot," in *2019 6th International conference on electrical engineering, computer science and informatics (EECSI)*. IEEE, 2019, pp. 106–111.

[13] S. Tiwari and S. Agarwal, "A shrewd artificial neural network-based hybrid model for pervasive stress detection of students using galvanic skin response and electrocardiogram signals," *Big Data*, vol. 9, no. 6, pp. 427–442, 2021.

[14] R. Chandra, S. Tiwari, A. Kumar, S. Agarwal, M. Syafrullah, and K. Adiyarta, "Autism spectrum disorder detection using autistic image dataset," in *2023 10th International Conference on Electrical Engineering, Computer Science and Informatics (EECSI)*. IEEE, 2023, pp. 54–59.

[15] S. Work, "Eeg psychiatric disorders dataset," https://www.kaggle.com/datasets/shashwatwork/eeg-psychiatric-disorders-dataset, 2023, accessed: 2025-01-09.